\documentclass[letterpaper, 10 pt, conference]{ieeeconf}  

\IEEEoverridecommandlockouts                              

\usepackage{url}

\usepackage{cite}
\usepackage{graphicx} 
\usepackage{color, soul}
\usepackage{array}
\usepackage{mathtools}
\usepackage{amsmath}
\usepackage{physics}
\usepackage{amsfonts}
\usepackage{subcaption}
\usepackage{gensymb}
\usepackage{varwidth}
\newcommand{\kpset}[1]{\mathbf{\kappa}_#1}
\newcommand{\basemap}{M_B}
\newcommand{\targetmap}{M_T}
\newcommand{\pseudomean}[1]{\widetilde{ \boldsymbol{\mu}}_{#1}}
\newcommand{\pseudovar}[1]{\widetilde{ \boldsymbol{\Sigma}}_{#1}}

\newcommand{\divhessian}[1]{
\begin{align}
    \mathbf{H}(\mathbf{#1}) = 
    \begin{bmatrix}
        \frac{\partial^2}{\partial x\partial x}\norm{{\mathbf{B}}(\mathbf{#1})} & \frac{\partial^2}{\partial x\partial y}\norm{{\mathbf{B}}(\mathbf{#1})} & \frac{\partial^2}{\partial x\partial z}\norm{{\mathbf{B}}(\mathbf{#1})} \\
        \frac{\partial^2}{\partial y\partial x}\norm{{\mathbf{B}}(\mathbf{#1})} & \frac{\partial^2}{\partial y\partial y}\norm{{\mathbf{B}}(\mathbf{#1})} & \frac{\partial^2}{\partial y\partial z}\norm{{\mathbf{B}}(\mathbf{#1})} \\
        \frac{\partial^2}{\partial z\partial x}\norm{{\mathbf{B}}(\mathbf{#1})} & \frac{\partial^2}{\partial z\partial y}\norm{{\mathbf{B}}(\mathbf{#1})} & \frac{\partial^2}{\partial z\partial z}\norm{{\mathbf{B}}(\mathbf{#1})} 
    \end{bmatrix}.
    \label{eq:div_hessian}
\end{align}
}

\title{\LARGE \bf Mag-Match: Magnetic Vector Field Features for \\Map Matching and Registration}

\author{William McDonald, Cedric Le Gentil, Jennifer Wakulicz and Teresa Vidal-Calleja
\thanks{All authors are with the Robotics Institute, University of Technology Sydney, Australia (e-mail: william.t.mcdonald@student.uts.edu.au, \{teresa.vidalcalleja,jennifer.wakulicz\}@uts.edu.au, cedric.legentil@uts.edu.au/@robotics.utias.utoronto.ca). %
This work was supported by the Australian Research Council Discovery Project under Grant DP210101336.}
}

\begin{document}
\maketitle

\begin{abstract}
Map matching and registration are essential tasks in robotics for localisation and integration of multi-session or multi-robot data. Traditional methods rely on cameras or LiDARs to capture visual or geometric information but struggle in challenging conditions like smoke or dust. Magnetometers, on the other hand, detect magnetic fields, revealing features invisible to other sensors and remaining robust in such environments. In this paper, we introduce \emph{Mag-Match}, a novel method for extracting and describing features in 3D magnetic vector field maps to register different maps of the same area. Our feature descriptor, based on higher-order derivatives of magnetic field maps, is invariant to global orientation, eliminating the need for gravity-aligned mapping. To obtain these higher-order derivatives map-wide given point-wise magnetometer data, we leverage a physics-informed Gaussian Process to perform efficient and recursive probabilistic inference of both the magnetic field and its derivatives.
We evaluate Mag-Match in simulated and real-world experiments against a SIFT-based approach, demonstrating accurate map-to-map, robot-to-map, and robot-to-robot transformations—even without initial gravitational alignment.
\end{abstract}

\section{Introduction}

Environmental awareness is the foundation for any robotic system to safely and successfully navigate throughout its environment, and often combines multiple perception modalities for increased capabilities \cite{robperception}. Tasks like map matching and registration can be found in re-localisation and loop closure in Simultaneous Localisation and Mapping (SLAM). Techniques used must be robust to changes in pose, as robots will most likely return to an area from a different angle. Therefore, recognising common areas measured from different perspectives is a critical property in matching algorithms.

Camera and LiDAR systems are widely used due to the type and abundance of data they can collect. However, cameras can suffer in low light conditions, be unable to distinguish between similar-looking objects, or not be permitted due to privacy restrictions, while LiDAR systems are often bulky and complex. Additionally, both suffer from adversarial conditions such as fog, dust, and heavy precipitations. In this paper, we propose using a magnetometer as magnetic fields are not affected by such conditions. These fields are commonly distorted by man-made structures, providing information about the environment that we use for mapping and feature extraction. The point measurements of a magnetometer further lend themselves naturally to data inference methods such as Gaussian Processes.

\begin{figure}[t]
    \centering
    \includegraphics[width=0.72\linewidth]{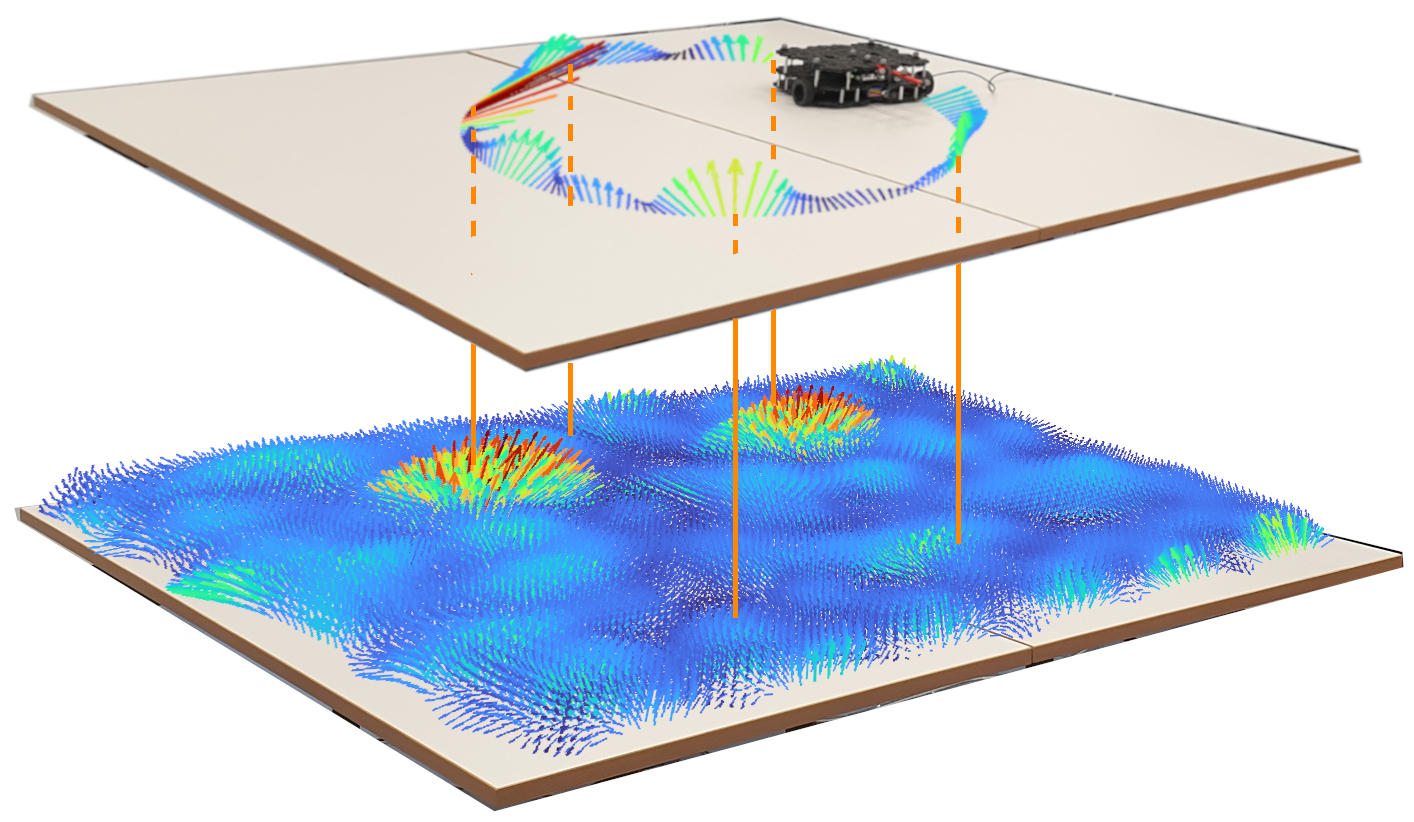}
    \caption{Mag-Match localising a robot to a previously mapped magnetic field region. Features from two GP magnetic vector field maps are detected and matched to enable registration.\vspace{-2mm}}
    \label{fig:full_overlay}
\end{figure}

Gaussian Process (GP) regression~\cite{gpforml} fits probabilistic functions to data by assessing correlations between data to estimate new values, where correlations are captured by user-defined kernel functions.  Works like~\cite{legentil2020gpgradientmaps,giubilato2022gpgmslam} have shown the potential for geometry-based loop-closure detection using GP representations in challenging environments. In this work, we leverage GP kernels~\cite{wahlstromlarge} with linear operators that enforce constraints given by laws of electromagnetic fields \cite{maxwell1865} to map the 3D magnetic vector field. Based on this representation, we present Mag-Match, a framework for keypoint detection and description for magnetic map registration. The keypoints correspond to high-curvature, low-variance points of the magnetic field and the descriptors are rotationally-invariant. This approach provides a new method of registration for applications such as loop closure, localisation, and multi-session or multi-robot mapping (see Fig.~\ref{fig:full_overlay}).

We evaluate Mag-Match on simulated and real-world environments with diverse magnetic fields to demonstrate effective salient feature detection, matching and multi-map transformation recovery. We benchmark our approach with a widely utilised detector and descriptor for 2D images using cross-sections of the vector field. This last approach requires a similar gravity alignment in the different maps to perform best. On the other hand, Mag-Match is agnostic to the global orientation of the maps. Further, we evaluate Mag-Match in a real environment from in-house and publicly available datasets.

\section{Related work}

Gaussian Processes have been widely utilised in magnetic field mapping to generate dense fields from sparse observations~\cite{wahlstromsmall,myICASSP,sarkka2011,magGP1,magGP2}. In~\cite{wahlstromsmall}, the authors use a kernel constrained by Maxwell's equations of electromagnetism~\cite{maxwell1865} to provide essential information to generate accurate magnetic \emph{vector} fields. Previously we proposed to use this physics constrained mapping for global localisation~\cite{myICASSP}. Here, we develop a feature detector and descriptor exploiting magnetic vector field properties and linear operations on GPs~\cite{sarkka2011}.

Mapping and navigation with magnetic fields have been well-explored. The authors in~\cite{basicmagmap} show magnetic mapping by gathering data in boustrophedon paths with linear interpolation for space-filling, but only in 2D layers. The work in~\cite{ambientSLAM} uses random motion exploration to gather data and model the three scalar components of the field separately, whereas we include component correlation. These correlations are used in~\cite{hexmaps} where large-scale maps are segmented into hexagonal domains. Extensions of prior methods have been shown in~\cite{footSLAM}, where the FootSLAM algorithm was enhanced with magnetic fields using a foot-mounted IMU with human motion analysis, improving 2D SLAM results. Other work exploring multi-sensor fusion includes~\cite{vmag}, where local variations of magnetic fields in indoor environments are used to complement visual information to improve localisation, highlighting magnetic sensing's handling of traditional sensor limitations. Although this work does not tackle localisation, it generates a 3D magnetic vector field in a recursive manner, lending itself nicely to future 3D SLAM and multi-sensor fusion solutions.

Features to match are a well-established method, especially for images. Lowe's widely used seminal work on the SIFT algorithm~\cite{ogSIFT} produces scale- and rotation-invariant descriptors for keypoints found in 2D images. Pixel gradient orientations quantify the local shape in a circle around a keypoint using HOG~\cite{hog} relative to its primary gradient orientation, which matches repeated occurrences in other images. Inspired by this work, we expand these ideas into 3D using a spherical descriptor space with local vector orientations and strengths, which we term Histograms of Oriented Vectors (HOV), aligning it to the Local Reference Frame (LRF) of the keypoint built from the local field structure to retain rotation-invariance.

Beyond image features, other works like \textit{Freetures}~\cite{millane2020freetures} match scalar fields using signed distance fields (SDFs) to extract features from the free space in map geometry. A spherical descriptor is built on the keypoints for rotational-invariance, also inspired by SIFT and HOG. The addition of this new feature space achieves good results over previous surface geometry-based matching methods FPFH~\cite{fpfh} and SHOT~\cite{shot}. In this work, we propose a similar descriptor, but exploit the use of GPs with linear operators to compute the Hessian analytically without relying on surrounding data for numerical calculation. Moreover, using vector fields over scalar fields also provides more information about the environment and is used to generate our LRFs.

\section{Methodology} \label{sec:methodology}

\begin{figure}
    \centering
    \includegraphics[width=1\linewidth]{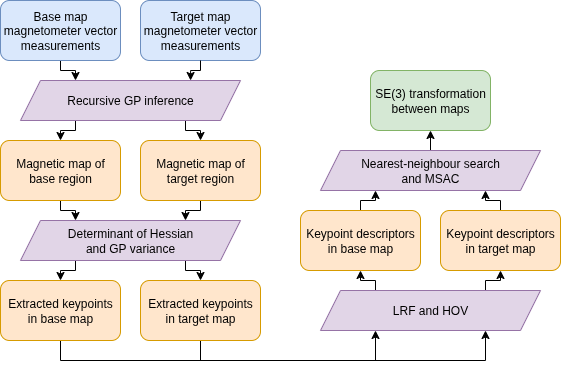}
    \caption{Stages of Mag-Match: recursive GP inference, feature extraction and description, and map matching.\vspace{-2mm}}
    \label{fig:pipeline}
\end{figure}

An overview of the proposed Mag-Match pipeline is shown in Fig.~\ref{fig:pipeline}, which aims to find the $SE(3)$ transformation between two sets of magnetometer measurements and their locations. Given three-axis magnetometer measurements, a magnetic vector field is recursively built using a divergence-free Gaussian Process.
This physics-informed GP generates a magnetic field map in a region bounded by the measurements for a base map \textit{$\basemap$} and a target map \textit{$\targetmap$}. The Determinant of Hessian (DoH) and GP variance are used to extract features that have high curvature and have been well-sampled by the measurements. The descriptors are generated in the keypoints' LRFs using HOV. Finally, these feature sets are registered using nearest-neighbour search and MSAC to recover the map-to-map transform, $T_{\basemap}^{\targetmap}$.

\subsection{3D Magnetic Vector Field Mapping with GPs}

GP regression is a kernel-based, data-driven, probabilistic interpolation technique~\cite{gpforml}, which can estimate data points between measurements, such as those on the trajectory of a magnetometer moving through a magnetic field. Let us consider the magnetic vector field $\mathbf{B}(\mathbf{x}) \in \mathbb{R}^3$ at locations $\mathbf{x} \in \mathbb{R}^3$ to be modelled with a zero-mean GP as
\begin{align}
    \mathbf{B}(\mathbf{x}) &\sim \mathcal{GP} \left( \mathbf{0}, \mathbf{K}(\mathbf{x}, \mathbf{x}')\right).
\end{align}
The kernel function $\mathbf{K} : \mathbb{R}^3 \times \mathbb{R}^3 \rightarrow \mathbb{R}^{3\times 3}$ captures the covariance between $\mathbf{B}$ at two locations $\mathbf{x}$ and $\mathbf{x}'$. 

Given a set of noisy magnetometer measurements $\mathbf{b}_i = \mathbf{B}(\mathbf{x}_i) + \boldsymbol{\epsilon}_i$ with $\boldsymbol{\epsilon}_i \sim \mathcal{N}(\mathbf{0},\boldsymbol{\Sigma}_z)$ taken at sensing locations $\{\mathbf{x}_{i}\}_{i=1}^N$, we wish to infer the magnetic field vector at an unseen location $\mathbf{x}^* \in \mathbb{R}^3$.
The multivariate Gaussian distribution of the GP model instantiated at the measurements and query point is given by
\begin{align}
    \begin{bmatrix}
    \mathbf{b} \\ \mathbf{B}(\mathbf{x}^*)
    \end{bmatrix}
    \sim \mathcal{N}
    \left(\mathbf{0},
    \begin{bmatrix}
    \mathbf{K}(\mathbf{X},\mathbf{X})+\mathbf{\Sigma}_z^2 & \mathbf{K}(\mathbf{X},\mathbf{x}^*)\\
    \mathbf{K}(\mathbf{x}^*,\mathbf{X}) & \mathbf{K}(\mathbf{x}^*,\mathbf{x}^*)
    \end{bmatrix}
    \right),
    \label{eq:multigaussian}
\end{align}
where $\mathbf{X} = [\mathbf{x}_1,\ldots,\mathbf{x}_N]^{\mathsf{T}} \in \mathbb{R}^{3\times N}$ is the array of measurement locations, $\mathbf{b} = [\mathbf{b}_1,\ldots,\mathbf{b}_N]^\mathsf{T}\in\mathbb{R}^{3\times N}$ the array of corresponding measurements, $\mathbf{K}(\mathbf{X},\mathbf{X}) \in \mathbb{R}^{3N\times 3N}$ is the Gram kernel covariance matrix of the measurement locations with $i$-$j$th $3\times 3$ block $[\mathbf{K}(\mathbf{X},\mathbf{X})]_{ij} = \mathbf{K}(\mathbf{x}_i,\mathbf{x}_j)$, and $\mathbf{K}(\mathbf{X},\mathbf{x}^*) = \mathbf{K}(\mathbf{x}^*,\mathbf{X})^\top$ the cross-covariance matrix between the query point and measurement locations.
By conditioning the distribution with respect to the measurements, the posterior mean $\boldsymbol{\mu}(\mathbf{x}^*)$ and covariance $\boldsymbol{\Sigma}(\mathbf{x}^*)$ at the new location can be obtained via
\begin{equation}
    \boldsymbol{\mu}(\mathbf{x}^*) = \mathbf{K}^{\mathsf{T}}_{\mathbf{X}}(\mathbf{x}^*)\left(\mathbf{K}_\mathbf{X}+\mathbf{\Sigma}_z^2 \right)^{-1} \mathbf{b},
    \label{eq:gpinfermean}
\end{equation}
\begin{equation}
    \boldsymbol{\Sigma}(\mathbf{x}^*) = \mathbf{K}(\mathbf{x}^*,\mathbf{x}^*) -
    \mathbf{K}^{\mathsf{T}}_{\mathbf{X}}(\mathbf{x}^*) \left(\mathbf{K}_{\mathbf{X}}+\mathbf{\Sigma}_z^2\right)^{-1} \mathbf{K}_{\mathbf{X}}(\mathbf{x}^*),
    \label{eq:gpinfervar}
\end{equation}
where we use the shorthand $\mathbf{K}_{\mathbf{X}}(\mathbf{x}^*) = \mathbf{K}(\mathbf{X},\mathbf{x}^*)$, $\mathbf{K}_{\mathbf{X}} = \mathbf{K}(\mathbf{X},\mathbf{X})$.

Furthermore, by careful choice of the kernel function $\mathbf{K}$, we can enforce the divergence-free nature of the magnetic \textbf{B}-field~\cite{maxwell1865} as a constraint during regression, improving the accuracy of inference. Linear operators applied to a GP produce another GP~\cite{sarkka2011}, and can be used to apply this constraint, as exploited in~\cite{linopsconstrained}, where the squared-exponential kernel is modified to become
\begin{align}
    \textbf{K}^B&(\mathbf{x}, \mathbf{x}') = \tfrac{\sigma^2_f}{l^2} \exp\left(- \tfrac{||\mathbf{x} - \mathbf{x}'||^2}{2l^2} \right) \cdot
    \nonumber \\
    &\left( \left( \tfrac{\mathbf{x} - \mathbf{x}'}{l} \right) \left( \tfrac{\mathbf{x} - \mathbf{x}'}{l} \right)^{\mathsf{T}} + \left( 2 - \tfrac{||\mathbf{x} - \mathbf{x}'||^2}{l^2} \right) I_{3} \right),
    \label{eq:divfreekernel}
\end{align}
with $\sigma_f^2$ a scaling factor and $l$ the lengthscale of the kernel. Using Eq.~\eqref{eq:gpinfermean} with $\mathbf{K}^B$ allows physics-informed inference of the magnetic vector \textbf{B}-field at any unseen locations $\mathbf{x}^*$.

The inversion of the Gram kernel matrix in Eqs.~\eqref{eq:gpinfermean}~and~\eqref{eq:gpinfervar} quickly causes standard GP regression to become computationally expensive as the number of measurements $N$ increases. Furthermore, robotic sensors commonly stream data over time, lending themselves to sequential GP updating that fuses new data points as they become available.
To this end, we employ a recursive sparse GP~\cite{rGP}. Here, a set of inducing points $\mathbf{U} = \{\mathbf{u}_i\}_{i=1}^U$ in the input domain are introduced, where $U << N$. Correlations between data are captured via the `pseudo-measurements' at this smaller set of points, making inference more efficient. Furthermore, a recursive update for the pseudo-measurements held at inducing points allows the \textit{full} GP belief to be updated online. The recursive procedure presented in~\cite{rGP} is summarised as follows.

Let $\widetilde{\boldsymbol{\mu}}$, $\widetilde{\boldsymbol{\Sigma}}$ be the mean and covariance of the pseudo-measurements or inducing values held at the inducing points. At $t=0$, a zero-mean prior $\pseudomean{0}=0$ is set with initial covariance $\pseudovar{0}= \mathbf{K}_{\mathbf{U}}=\mathbf{K}^B(\mathbf{U},\mathbf{U})$. Given measurements up to time $t$, a prediction $\widehat{\mathbf{b}}_{t+1}$ at location $\mathbf{x}_{t+1}$ is given as
\begin{equation}
\begin{array}{l}
    \widehat{\mathbf{b}}_{t+1} = \pseudomean{t} \\
    \Sigma^{\mathbf{b}\mathbf{b}}_{t+1} = \widetilde{\mathbf{K}}_*(\mathbf{x}_t, \mathbf{x}_t) + \sigma^2_\epsilon \\
    \Sigma^{\mathbf{b}\mathbf{U}}_{t+1} = \mathbf{K}^B(\mathbf{x}_t,\mathbf{U}) \mathbf{K}_\mathbf{U}^{-1} \pseudovar{t}.
\end{array}\label{eq:rGP:predict}
\end{equation}
$\widetilde{\mathbf{K}}_*$ is a sparse kernel, where we use Subset of Regressors (SoR)~\cite{sparseGPs}, expressed as: 
\begin{align*}
    \widetilde{\mathbf{K}}_{SoR}(\mathbf{x}_t, \mathbf{x}_t') = \mathbf{K}^B(\mathbf{x}_t,\mathbf{U}) \mathbf{K}_{\mathbf{U}}^{-1} \pseudovar{t} \mathbf{K}_{\mathbf{U}}^{-1} \mathbf{K}^B(\mathbf{U},\mathbf{x}'_t).
\end{align*}

When measurement $\mathbf{b}_{t+1}$ becomes available, the belief at $\mathbf{M}$ can be updated using the predictions in Eq.~\eqref{eq:rGP:predict},
\begin{equation}
\begin{array}{l}
    \pseudomean{t+1} = \pseudomean{t} + \Sigma^{\mathbf{b}\mathbf{U}}_{t+1}(\Sigma^{\mathbf{b}\mathbf{b}}_{t+1})^{-1}(\mathbf{b}_{t+1} - \widehat{\mathbf{b}}_{t+1}) \\
    \pseudovar{t+1} = \pseudovar{t} - \Sigma^{\mathbf{b}\mathbf{U}}_{t+1}(\Sigma^{\mathbf{b}\mathbf{b}}_{t+1})^{-1}(\Sigma^{\mathbf{b}\mathbf{U}}_{t+1})^{\mathsf{T}}.
\end{array}
\end{equation}

The belief held at the inducing points can then be used in full GP regression, where inference at any query point $\mathbf{x}^*$ is recovered with
\begin{equation}
\begin{array}{l}
    \boldsymbol{\mu}(\mathbf{x}^*) =\mathbf{K}^B(\mathbf{x}^*,\mathbf{U}) \mathbf{K}_{\mathbf{U}}^{-1} \pseudomean{t}
    \\
    \boldsymbol{\Sigma}(\mathbf{x}^*) = \mathbf{K}^B(\mathbf{x}^*,\mathbf{U}) \mathbf{K}_{\mathbf{U}}^{-1}\pseudovar{t}\mathbf{K}_{\mathbf{U}}^{-1}\mathbf{K}^B(\mathbf{U},\mathbf{x}^*). \label{eq:rgp:inference}
\end{array}
\end{equation}

Importantly, the inversion $\mathbf{K}_{\mathbf{U}}^{-1}$ is performed only once during initialisation. Placing $\mathbf{U}$ on a regular grid can also lead to structure in the kernel, further speeding up inversion and alleviating the computational cost of querying the GP.

\subsection{Keypoint selection and description}

\begin{figure}
    \centering
    \includegraphics[width=1\linewidth]{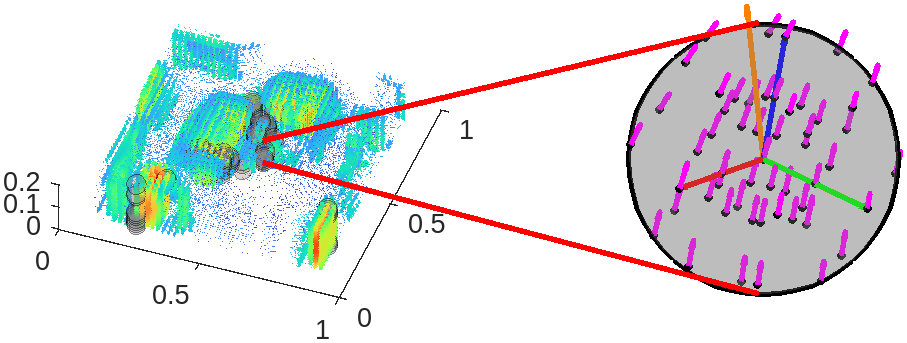}
    \caption{Design of feature descriptor sphere. Left shows an inferred magnetic vector field with detected keypoints in transparent grey. Right shows a magnified keypoint. The $x$-$y$-$z$ axes of the LRF are shown in RGB, respectively, the base vector in orange and the set of subpoints, $\mathbf{S}_{\kpset{i}}$, in black with their describing vectors, $\mathbf{V}_{\kpset{i}}$, in magenta.\vspace{-2mm}}
    \label{fig:kp_field_with_zoom}
\end{figure}

To identify keypoints and corresponding descriptors, we first assess the curvature of the magnetic field at a set of inference points $\mathbf{X}^*$ by evaluating the Hessian of the norm of the magnetic field. Then, a subset of $\mathbf{X}^*$ is selected as keypoints and their corresponding LRFs are defined using the shape of the base vectors in $\mathbf{B}(\mathbf{X}^*)$ from Eq.~\eqref{eq:gpinfermean}. Finally, the magnitudes and orientations of the nearby field vectors within a radius in this LRF are used to form a descriptor vector using HOV.
An example magnetic field with keypoints is shown in Fig.~\ref{fig:kp_field_with_zoom}, with a magnified view of a keypoint and its spherical descriptor region. The details of keypoint selection and descriptor construction are given below.

\subsubsection{Determinant of Hessian (DoH) and GP Variance}

As the shape of a magnetic vector field is generally constant and indistinct except near a magnetic source, regions of high curvature provide the best keypoint locations. To identify such locations, we employ the DoH metric on the \textit{norm} of the magnetic field.
Fig.~\ref{fig:office_room_doh_heat_map}) shows an example DoH map of a slice through the region in Fig.~\ref{fig:office_room_model}), where the edges and corners of objects can be identified. The Determinant of Hessian metric is defined as
\begin{equation}
    DoH(\mathbf{x}^*) = \det(\mathbf{H}(\mathbf{x}^*)),
\end{equation}
where 
\begin{small}
    \divhessian{x^*}
\end{small}

The general form for the elements of $\mathbf{H}(\mathbf{x}^*)$ is given by
\begin{equation}
\begin{array}{c}
    [\mathbf{H}(\mathbf{x}^*)]_{ij} = \frac{\partial {\mathbf{B}}(\mathbf{x}^*)}{\partial i} \frac{\partial {\mathbf{B}^{\mathsf{T}}}(\mathbf{x}^*)}{\partial j} \cdot \left(\frac{1}{\norm{{\mathbf{B}}(\mathbf{x}^*)}} - \frac{{\mathbf{B}}(\mathbf{x}^*){\mathbf{B}}^{\mathsf{T}}(\mathbf{x}^*)}{\norm{{\mathbf{B}}(\mathbf{x}^*)}^3}\right) + \\ \left(\frac{{\mathbf{B}}(\mathbf{x}^*)}{\norm{{\mathbf{B}}(\mathbf{x}^*)}}\right) \left(\frac{\partial^2\norm{{\mathbf{B}}(\mathbf{x}^*)}}{\partial i \partial j}\right).
    \label{hessian_component_general}
\end{array}
\end{equation}

Application of linear operators to the divergence-free kernel in Eq. \eqref{eq:divfreekernel} enables analytic evaluation of the DoH, avoiding the numerical approximations found in all prior works such as~\cite{millane2020freetures}. The operators $\mathcal{L}_{PD1}=\frac{\partial}{\partial i}$ and $\mathcal{L}_{PD2}=\frac{\partial^2}{\partial i \partial j}$ produce a set of three first-order and nine second-order partial derivative kernels (six of the latter due to symmetry). 

To identify keypoints, the DoH is first calculated for all points in $\mathbf{X}^*$. The mean of these values is used as the minimum threshold to filter out low-curvature areas. Then, points with high variance (Eq.~\eqref{eq:gpinfervar}) are removed to limit the selection to areas of high confidence. The remaining points $\kappa = \{\kappa_i\}_{i=1}^{K}$ are considered keypoints.

\subsubsection{Local Reference Frame (LRF)}

For each keypoint, the $z$-axis is set as the normalised vector $\hat{\mathbf{B}(\mathbf{x}^*_{\kappa_i})}$ inferred at the keypoint. To align the $x$-axis to the surrounding vector field, a spherical region with radius 4 times the inference point spacing defines the keypoint support, $\boldsymbol{\Omega}_{\kappa_i}$. The inferred field vectors $\mathbf{B}(\mathbf{X}^*)$ within this support are projected onto the plane perpendicular to the $z$-axis. Gaussian weighting places more importance on closer vectors. The vector sum produces one vector that is aligned to the local shape of the surrounding field, which is then normalised and assigned as the $x$-axis of the LRF. Finally, the $y$-axis is chosen to give a complete right-handed coordinate system.
These axes form the basis vectors defining the LRF $\mathcal{F}_{\kpset{i}}$ for keypoint ${\kappa_i}$.

\subsubsection{Histograms of Oriented Vectors (HOV)}

To capture the shape of the local vector field around a keypoint, the strengths and orientations of the vectors within a keypoint's support are assessed using our variation of HOG \cite{hog}, which we refer to as the Histograms of Oriented Vectors (HOV).

For each keypoint, support vectors $\mathbf{B}(\mathbf{X}^*)$ within $\boldsymbol{\Omega}_{\kpset{i}}$, are inferred in the map frame and transformed to the LRF using $T^{map}_{\kpset{i}}$, which is the transpose of $T_{map}^{\kpset{i}}$ that defines $\mathcal{F}_{\kpset{i}}$. HOV captures the orientations and strengths of these support vectors. Their azimuth ($\pm180\,\degree$) and elevation ($\pm90\,\degree$) angles in the LRF are calculated and placed into bins of size $18\,\degree$ each, producing a 20-dimensional azimuth vector and a 10-dimensional elevation vector. The $x$, $y$, and $z$ component bin sizes are chosen based on the scale and units of the data at hand. In this work, the range is set to $\pm100\,\mathrm{G}$ and the bins sizes to $10\,\mathrm{G}$. Values existing outside this range are discarded. This produces a 20-dimensional vector for each component. These are combined with the angle histograms to form a final 90-dimensional descriptor for each keypoint.

\subsection{Map registration}

Consider now two sets of keypoints, $\kappa^{\basemap}$ and $\kappa^{\targetmap}$, in a base map $\basemap$ and a target map $\targetmap$, respectively. To estimate $T_{\basemap}^{\targetmap} \in SE(3)$, a nearest-neighbour search in the HOV descriptor space is used to find correspondences between $\kappa^{\basemap}$ and $\kappa^{\targetmap}$ and a threshold is applied to the matching distance to define true correspondences. This produces the set of all correspondences, $\mathcal{C}=\{c_1, c_2,\cdots,c_N\}$, where $c_i$ contains a correspondence between $\kappa^{\basemap}$ and $\kappa^{\targetmap}$.

We introduce a modified version of M-estimator SAmple Consensus (MSAC) registration~\cite{msac} that leverages vector information to estimate $T_{\basemap}^{\targetmap}$ from $\mathcal{C}$. Traditional MSAC evaluates the Euclidean distances between correspondences to identify inliers. Our modified MSAC additionally includes evaluation of the cross product between vectors in corresponding pairs, which is minimised for vectors that align well. As a final fitness check, the estimated transform is applied to the samples in $\targetmap$ to align their vectors with those in $\basemap$. If the average norm of the cross product between aligned vectors is below a user-defined threshold, we consider the transformation recovered.

\begin{figure}[t]
    \centering
    \begin{subfigure}[t]{0.4\linewidth}
        \centering
        \includegraphics[width=1\linewidth]{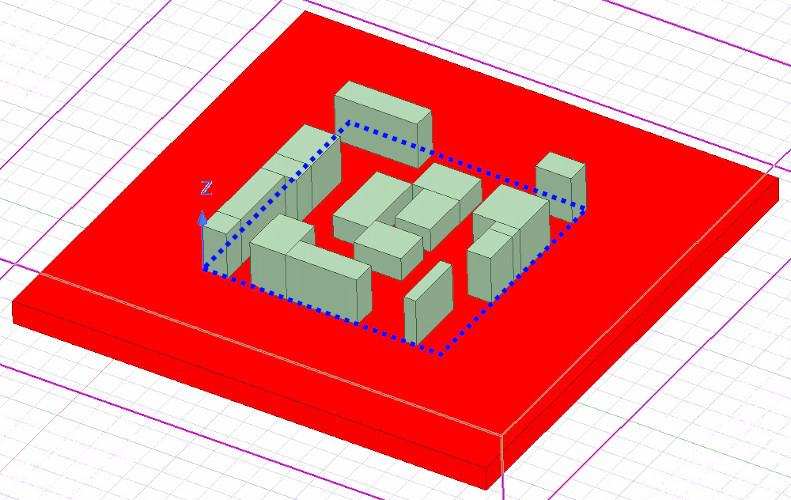}
        \caption{3D model}
        \label{fig:office_room_model}
    \end{subfigure}
    \begin{subfigure}[t]{0.36\linewidth}
        \centering
        \includegraphics[width=0.8\linewidth]{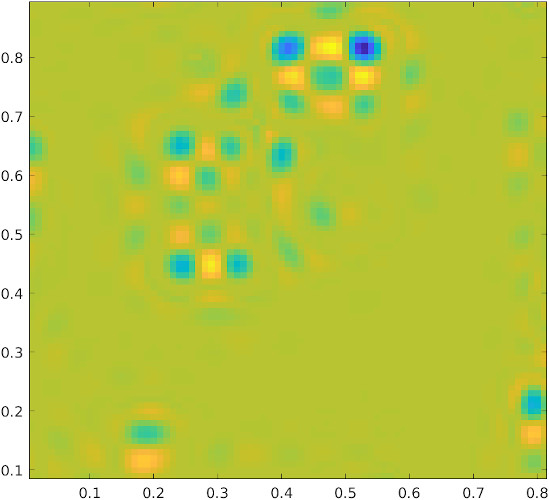}
        \caption{DoH heatmap at $z=0.02m$}
        \label{fig:office_room_doh_heat_map}
    \end{subfigure}
    \caption{Example simulation environment. The dashed region in (a) shows the bounds of the DoH heatmap shown in (b). It identifies the corners and edges of some of the obstacles at the sensor height. This is expected as sharp features of objects distort magnetic fields more than flat surfaces.}
    \label{fig:office_room}
\end{figure}

\begin{figure}[t]
    \centering
    \begin{subfigure}[t]{0.445\linewidth}
        \centering
        \includegraphics[width=\linewidth]{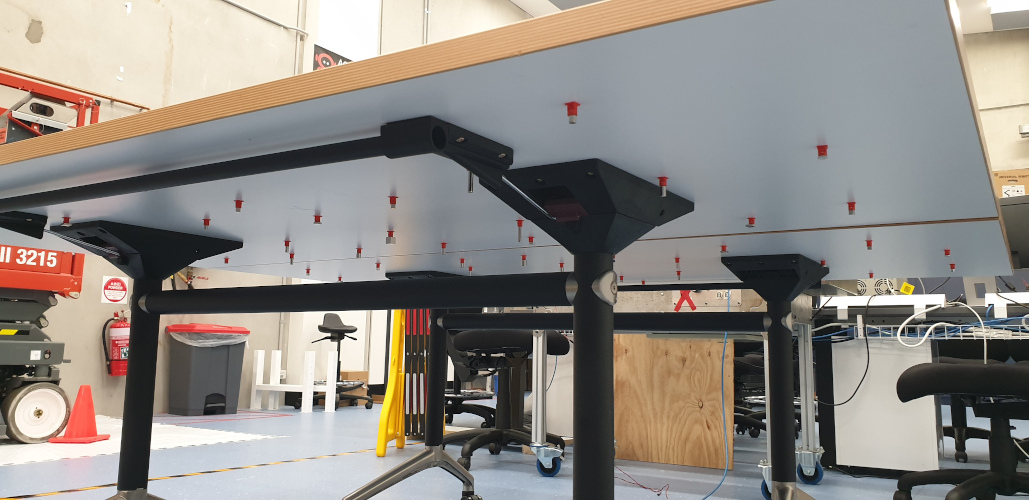}
        \caption{}
        \label{fig:vicon_setup}
    \end{subfigure}
    \begin{subfigure}[t]{0.54\linewidth}
        \centering
        \includegraphics[width=\linewidth]{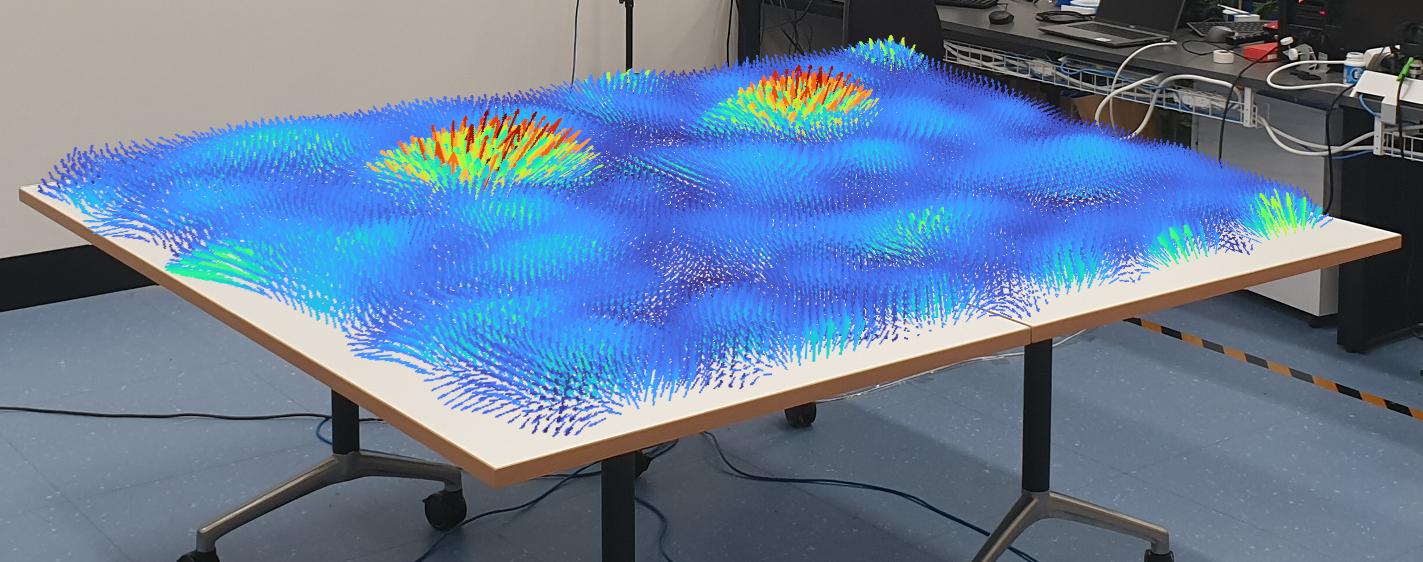}
        \caption{}
        \label{fig:vicon_heatmap}
    \end{subfigure}
    \caption{UTS Tech Lab experiment: (a) shows the location of the magnets affixed underneath the table, and (b) the superimposed inferred magnetic vector field.\vspace{-2mm}}
    \label{fig:vicon}
\end{figure}

\section{Experiments}

We evaluate Mag-Match in simulation and real-world environments. Robot trajectories and hand-mapped datasets to perform map-to-map, robot-to-map and robot-to-robot registration. For large mapping datasets, we define $\mathbf{X}^*$ on a regular grid bounded by the measurements. For robot trajectories, $\mathbf{X}^*$ is defined in a small radius along the path.
We implement a simple SIFT-based method for comparison. First, layers of vector data are inferred with the GP, from which image-like heat maps are created from vector norms. Then, SIFT is used on these heat maps for feature extraction. 

\subsection{Datasets}
We evaluate Mag-Match on a series of datasets, simulated and real, detailed below.
\subsubsection{ANSYS Simulation}

A simulation environment is created in ANSYS, which uses Finite Element Analysis to calculate the \textbf{B}-field. A space with various structures generating magnetic fields of different strengths is shown in Fig.~\ref{fig:office_room_model}), and three coordinate systems are defined: a base map (with no transformation from the world frame), and two target maps, one with a difference of $+30\degree$ in the $z$-axis and one with $-30\degree$ in the $x$-axis (non-gravity aligned). The sampling locations are uniformly distributed in the region for each coordinate system, and random sensor noise with $\sigma=0.01$T is added. We perform a 100-sample Monte Carlo analysis varying the input noise and find the RMS error and standard deviation.

\begin{figure}
    \centering
    ${\vcenter{\hbox{\includegraphics[width=0.5\linewidth]{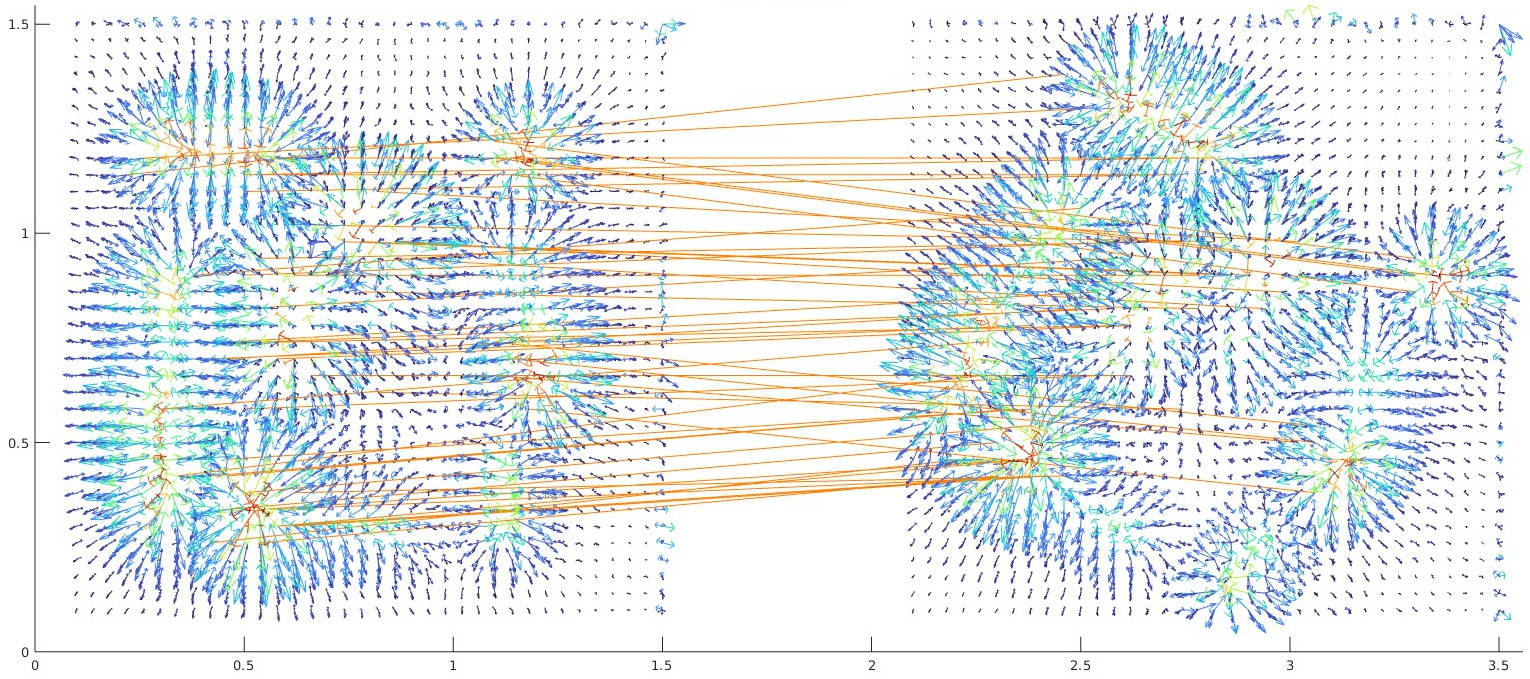}}}}$
    ${\vcenter{\hbox{$\Longrightarrow$}}}$
    ${\vcenter{\hbox{\includegraphics[width=0.4\linewidth]{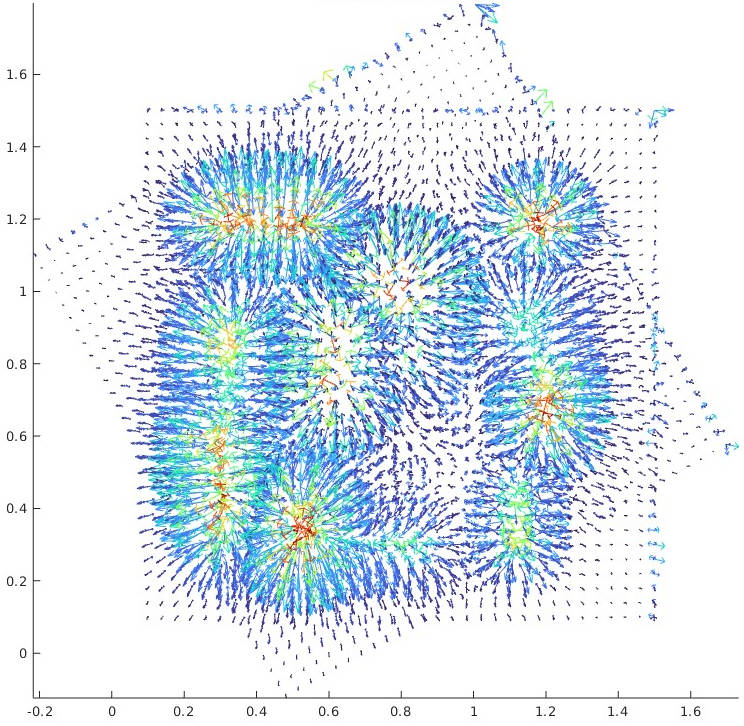}}}}$
    \caption{Simulation environment: registration and alignment between base and target map with $+30\degree$ offset in the $z$-axis. Orange lines indicate matched keypoints used for the gravity-aligned transformation recovery.}
    \label{fig:ansys_matching_+30z}
\end{figure}

\begin{figure}
    \centering
    ${\vcenter{\hbox{\includegraphics[width=0.5\linewidth]{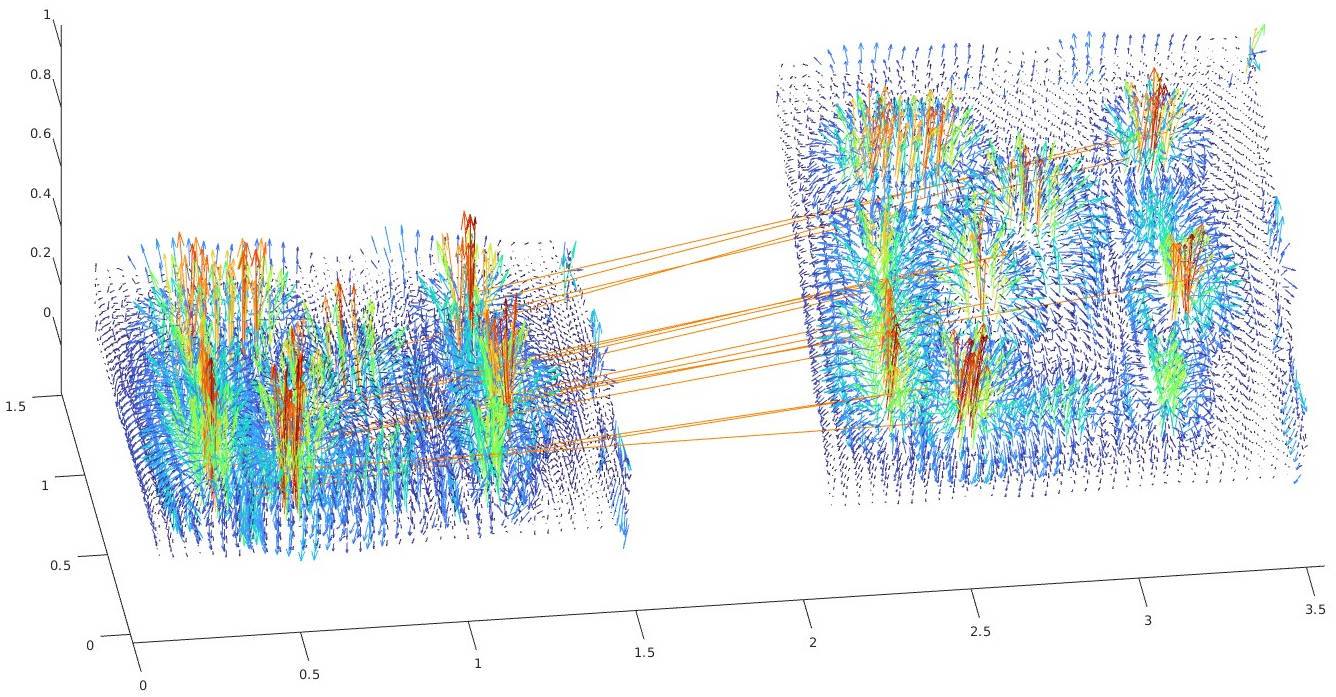}}}}$
    ${\vcenter{\hbox{$\Longrightarrow$}}}$
    ${\vcenter{\hbox{\includegraphics[width=0.4\linewidth]{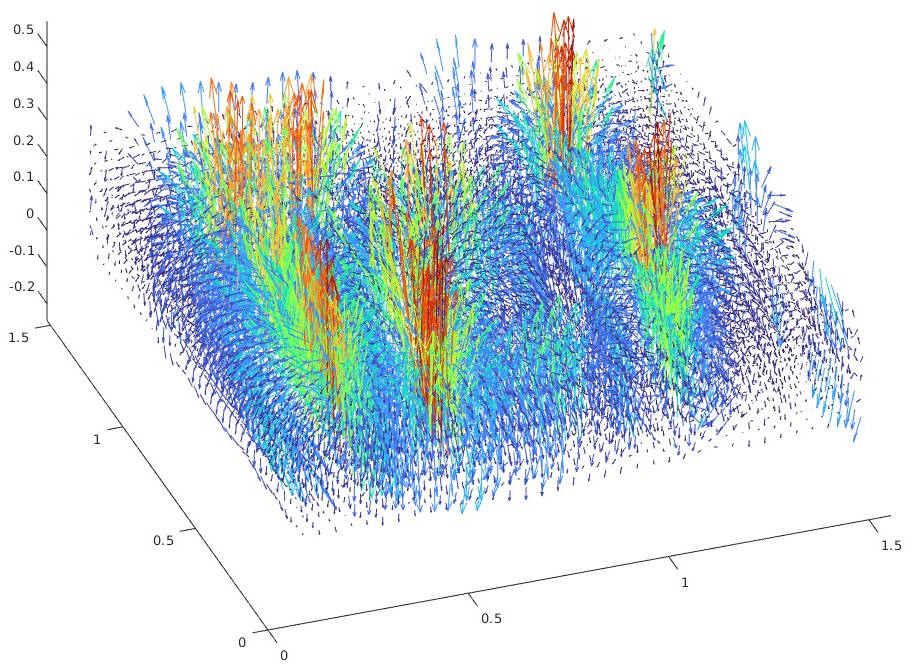}}}}$
    \caption{Simulation environment: registration and alignment between base and target map with $-30\degree$ offset in the $x$-axis. Orange lines indicate matched keypoints used for the 3D transformation recovery.\vspace{-2mm}}
    \label{fig:ansys_matching_-30x}
\end{figure}

\subsubsection{UTS Tech Lab}

Both map-to-map and robot-to-map registration are demonstrated with datasets collected at UTS Tech Lab. A $3$m$^2$ table surface with varying-strength magnets affixed underneath provides a diverse magnetic field, shown in Fig.~\ref{fig:vicon}).
Measurements are collected
using a calibrated Microstrain 3DM-GX5 magnetometer. A Vicon optical tracking system and the IMU-pose spatiotemporal extrinsic calibration proposed in \cite{legentil2023calib} is used to acquire sensor poses and align measurements to the map frame.

For map-to-map registration, two maps are formed from measurements collected along vertical and horizontal boustrophedon paths. The second map uses the initial sensor pose as its reference frame and is rotated by $20\degree$ in both $y$ and $z$ relative to the first. For map-to-robot, a differential-drive robot followed circular trajectories near the centre of the table. Importantly, the fixed height sensor in this setup only gathered measurements at approximately $0.05$m above the table, limiting the region in which the field can reliably be inferred using the GP.

\begin{figure}
    \centering
    \begin{subfigure}[t]{0.5\linewidth}
        ${\vcenter{\hbox{\includegraphics[width=\textwidth]{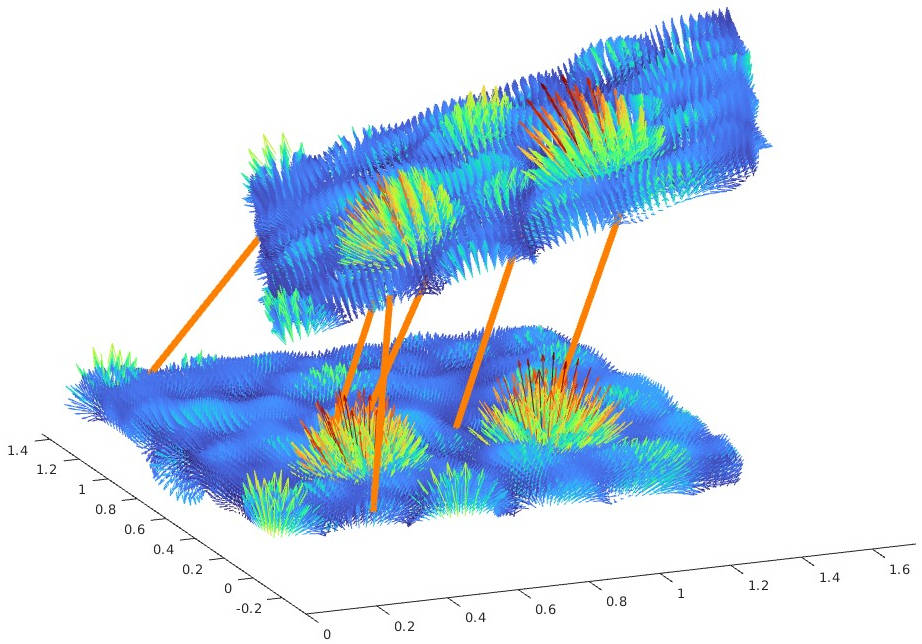}}}}$
        \label{fig:vicon_matched}
    \end{subfigure}
    \begin{subfigure}[t]{0.4\linewidth}
        ${\vcenter{\hbox{\includegraphics[width=\textwidth]{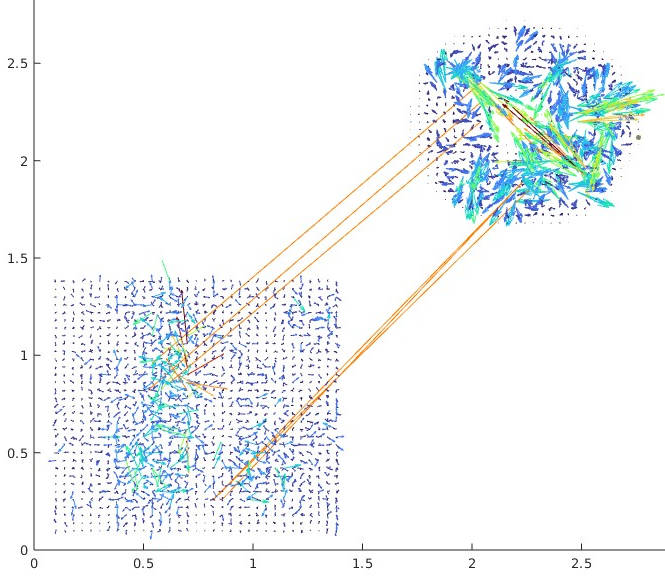}}}}$
        \label{fig:map_robot_matched}
    \end{subfigure}
    \\
    ${\vcenter{\hbox{$\downarrow$}}}$\hspace{3.5cm}
    ${\vcenter{\hbox{$\downarrow$}}}$
    \\
    \begin{subfigure}[t]{0.5\linewidth}
        ${\vcenter{\hbox{\includegraphics[width=\textwidth]{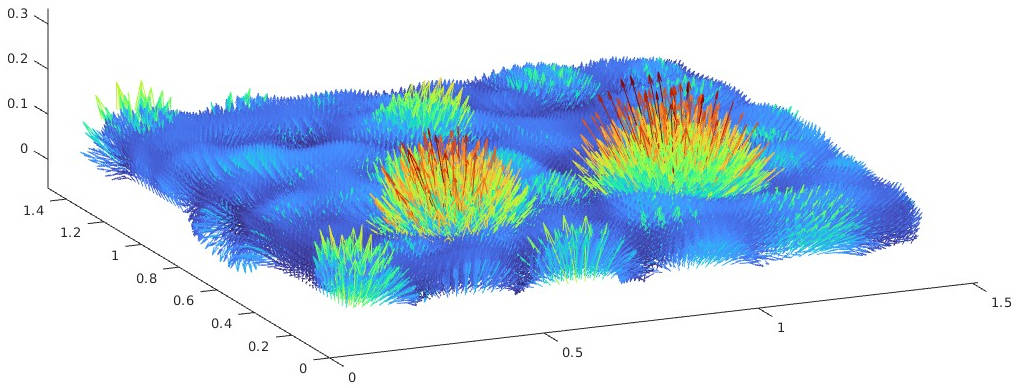}}}}$
        \caption{}
        \label{fig:vicon_aligned}
    \end{subfigure}
    \begin{subfigure}[t]{0.3\linewidth}
        ${\vcenter{\hbox{\includegraphics[width=\textwidth]{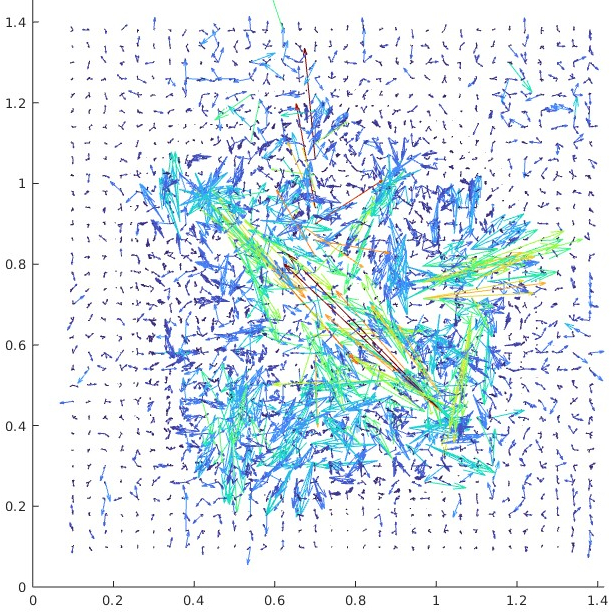}}}}$
        \caption{}
        \label{fig:map_robot_aligned}
    \end{subfigure}
    \caption{Registration and alignment of (a) map-to-map and (b) robot-to-map scenarios using Mag-Match in UTS Tech Lab experiment. Lines indicate matched keypoints used for transformation recovery.}
    \label{fig:vicon_matching}
\end{figure}

\subsubsection{KI Building dataset}

We demonstrate robot-to-robot registration on the publicly available dataset in~\cite{magposegraphslam}. In this work, magnetometer and wheel encoder data are given as inputs to a pose graph SLAM algorithm. One dataset -- the KI Building -- contains exploration of one region twice, allowing for robot-to-robot registration. Mag-Match and SIFT are run on each of these trajectories to evaluate each in a real environment. Here, the SLAM-corrected poses are used as the measurement locations. The limited coverage of the trajectories is challenging for feature extraction.

\begin{figure}
    \centering
    \begin{subfigure}[t]{0.445\linewidth}
        \centering
        \includegraphics[width=\linewidth]{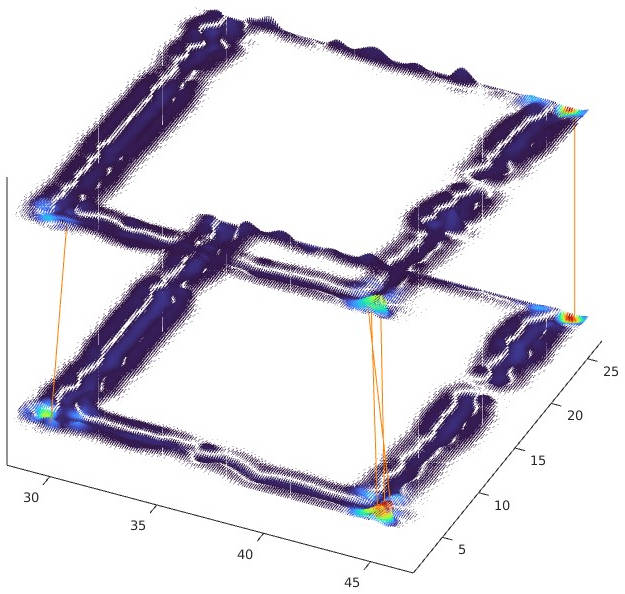}
        \caption{Mag-Match}
        \label{fig:poseslam_gp}
    \end{subfigure}
    \begin{subfigure}[t]{0.445\linewidth}
        \centering
        \includegraphics[width=\linewidth]{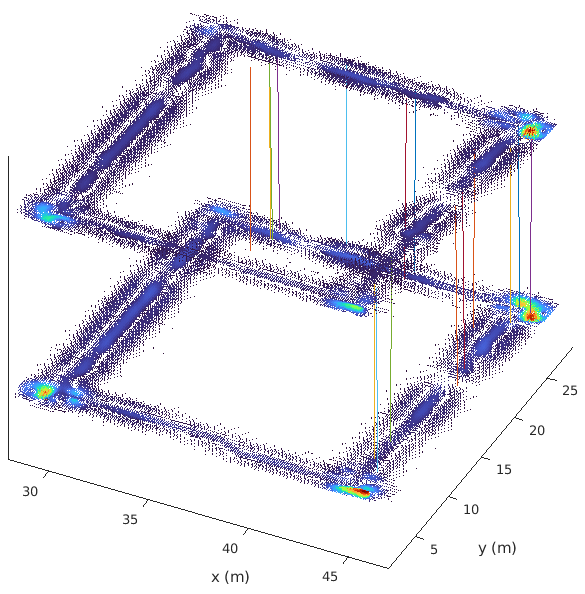}
        \caption{SIFT}
        \label{fig:poseslam_sift}
    \end{subfigure}
    \caption{Registration of robot-to-robot trajectories in KI Building dataset~\cite{magposegraphslam} using (a) Mag-Match and (b) SIFT. Lines indicate matched keypoints used for transformation recovery.\vspace{-2mm}}
    \label{fig:posegraph_dataset}
\end{figure}

\begin{table}[t]
    \centering
        \caption{Transformation recovery results. Values are RMSE pose errors from ground-truth transformation. For ANSYS simulations, the error is produced from varying the input noise. For the real-world experiments, the error is produced for repeated iterations of MSAC, which can produce different results due to its randomised nature~\cite{msac}.}
    \def\colsize{2.2cm}
    \def\colsizeh{2.5cm}
    \newcolumntype{C}[1]{>{\centering\let\newline\\\arraybackslash\hspace{0pt}}m{#1}}
        \begin{tabular}{C{\colsize}|C{\colsizeh}|C{\colsizeh}}
        Experiment & Mag-Match & SIFT\\
        \hline\hline
         Ansys ($+30\degree z$, gravity aligned) & 0.0816\,m $\pm$ 0.0299 3.2316\,\degree $\pm$ 1.8319 & \textbf{0.0198\,m $\pm$ 0.0920 2.0361\,\degree $\pm$ 0.8580}\\
         \hline
         Ansys ($-30\degree z$, non-gravity aligned) & \textbf{0.0194\,m $\pm$ 0.0036 0.9237\,\degree $\pm$ 0.4023} & 0.1416\,m $\pm$ 0.0646 10.3866\,\degree $\pm$ 7.6761 \\
         \hline
         UTS Tech Lab map-to-map (non-gravity aligned) & \textbf{0.0391\,m $\pm$ 0.0147 3.8719\,\degree $\pm$ 1.0645} & 0.1396\,m $\pm$ 0.0773 9.6106\,\degree $\pm$ 1.577\\
         \hline
         KI Building~\cite{magposegraphslam} map-to-map & \textbf{0.1784\,m $\pm$ 0.0836} 0.2688\,\degree $\pm$ 0.1076 & 0.2967\,m $\pm$ 0.3253 \textbf{0.2313\,\degree $\pm$ 0.2764}\\
         \hline
        \end{tabular}
    \label{tab:results}
\end{table}

\subsection{Results Discussion}

The experiments highlight some key advantages of Mag-Match (see Table~\ref{tab:results}). The simulation results are shown in Figs.~\ref{fig:ansys_matching_+30z} and~\ref{fig:ansys_matching_-30x}, where the orange links indicate matched keypoints used for transformation recovery. Both perform well in the gravity-aligned case, where SIFT edges out ahead thanks to consistent layer positions. However, in the non-gravity-aligned case, Mag-Match performs best, where the rotation-invariance of the keypoint descriptor identifies common keypoints despite the orientation difference. The dense training data allows SIFT to find common areas as well, but as layers used for feature extraction are not aligned across maps, it does not recover the transformation completely.

Results for the Tech Lab experiment are shown in Fig.~\ref{fig:vicon_matching}. In both, the majority of keypoints are detected around the two strongest magnets. This is sufficient for registration, as the physics-informed GP can infer changes in curvature due to the differences in the surrounding magnetic field. The non-gravity alignment in the map-to-map experiment means the features in each image-like heatmap used in the SIFT method have different surrounding pixels to build their histograms from, leading to different descriptors. Mag-Match, however, finds the same surrounding vector information in both cases thanks to its 3D descriptor space. Fig.~\ref{fig:map_robot_aligned}) shows Mag-Match localising the robot to the map. The results here were less consistent due to the limited information available in the robot's measurements to form useful keypoints. Despite this, Mag-Match was able to recover the transformation. An example of a successful match is shown in Fig.~\ref{fig:full_overlay}.

The KI Building dataset from~\cite{magposegraphslam} shows similar performance from both methods. Each identified keypoints at the corners where the magnetic field was strongest, in addition to those on the straight paths. In this experiment, the SLAM-corrected points were used, as the odometry points were significantly askew in the available dataset, so both methods had a well-correlated dataset to extract features from.

\section{Conclusion}

In this paper we proposed Mag-Match, a 3D magnetic vector field keypoint detector and descriptor. Our detector finds keypoints in regions of high curvature by leveraging a divergence-free recursive GP to infer higher-order magnetic field derivatives and corresponding variances at any location in the domain.
The GP further enabled establishment of local reference frames and keypoint descriptors based on inferred vectors to ensure consistent orientation over different initial poses. Evaluation demonstrated that Mag-Match outperformed SIFT where
an initial alignment to gravity is absent. In future work, we aim to develop a full SLAM pipeline leveraging the proposed magnetic features and explore the use of learning-based methods for feature description and matching.

 %
%
%
\bibliographystyle{IEEEbib}
\bibliography{library}

\begin{thebibliography}{10}

\bibitem{robperception}
Cristiano Premebida, Rares Ambrus, and Zoltan-Csaba Marton,
\newblock ``Intelligent robotic perception systems,''
\newblock in {\em Applications of Mobile Robots}. IntechOpen London, UK, 2018.

\bibitem{gpforml}
Carl~Edward Rasmussen and Christopher K.~I. Williams,
\newblock {\em {Gaussian Processes for Machine Learning}},
\newblock The MIT Press, 11 2005.

\bibitem{legentil2020gpgradientmaps}
Cedric Le~Gentil, Mallikarjuna Vayugundla, Riccardo Giubilato, Wolfgang Stürzl, Teresa Vidal-Calleja, and Rudolph Triebel,
\newblock ``Gaussian process gradient maps for loop-closure detection in unstructured planetary environments,''
\newblock in {\em 2020 IEEE/RSJ International Conference on Intelligent Robots and Systems (IROS)}, 2020, pp. 1895--1902.

\bibitem{giubilato2022gpgmslam}
Riccardo Giubilato, Cedric Le~Gentil, Mallikarjuna Vayugundla, Martin Schuster, Teresa Vidal-Calleja, and Rudolph Triebel,
\newblock ``Gpgm-slam: a robust slam system for unstructured planetary environments with gaussian process gradient maps,''
\newblock {\em Field Robotics}, vol. 2, pp. 1721--1753, 2022.

\bibitem{wahlstromlarge}
Niklas Wahlström,
\newblock {\em Modeling of Magnetic Fields and Extended Objects for Localization Applications},
\newblock Ph.D. thesis, Linköping University, 2015.

\bibitem{maxwell1865}
James~Clerk Maxwell,
\newblock ``A dynamical theory of the electromagnetic field,''
\newblock {\em Philosophical Transactions of the Royal Society of London}, vol. 155, pp. 459--512, 1865.

\bibitem{wahlstromsmall}
Niklas Wahlström, Manon Kok, Thomas~B. Schön, and Fredrik Gustafsson,
\newblock ``Modeling magnetic fields using gaussian processes,''
\newblock in {\em 2013 IEEE International Conference on Acoustics, Speech and Signal Processing}, 2013, pp. 3522--3526.

\bibitem{myICASSP}
William McDonald, Cedric Le~Gentil, and Teresa Vidal-Calleja,
\newblock ``Global localisation in continuous magnetic vector fields using gaussian processes,''
\newblock in {\em 2023 IEEE International Conference on Acoustics, Speech and Signal Processing (ICASSP)}, 2023, pp. 1--5.

\bibitem{sarkka2011}
Simo S{\"a}rkk{\"a},
\newblock ``Linear operators and stochastic partial differential equations in gaussian process regression,''
\newblock in {\em Artificial Neural Networks and Machine Learning -- ICANN 2011}, Timo Honkela, W{\l}odzis{\l}aw Duch, Mark Girolami, and Samuel Kaski, Eds., Berlin, Heidelberg, 2011, pp. 151--158, Springer Berlin Heidelberg.

\bibitem{magGP1}
Naoki Akai and Koichi Ozaki,
\newblock ``3d magnetic field mapping in large-scale indoor environment using measurement robot and gaussian processes,''
\newblock in {\em 2017 International Conference on Indoor Positioning and Indoor Navigation (IPIN)}, 2017, pp. 1--7.

\bibitem{magGP2}
Naoki Akai and Koichi Ozaki,
\newblock ``Gaussian processes for magnetic map-based localization in large-scale indoor environments,''
\newblock in {\em 2015 IEEE/RSJ International Conference on Intelligent Robots and Systems (IROS)}, 2015, pp. 4459--4464.

\bibitem{basicmagmap}
Etienne Le~Grand and Sebastian Thrun,
\newblock ``3-axis magnetic field mapping and fusion for indoor localization,''
\newblock in {\em 2012 IEEE International Conference on Multisensor Fusion and Integration for Intelligent Systems (MFI)}, 2012, pp. 358--364.

\bibitem{ambientSLAM}
Ilari Vallivaara, Janne Haverinen, Anssi Kemppainen, and Juha Röning,
\newblock ``Simultaneous localization and mapping using ambient magnetic field,''
\newblock in {\em 2010 IEEE Conference on Multisensor Fusion and Integration}, 2010, pp. 14--19.

\bibitem{hexmaps}
Manon Kok and Arno Solin,
\newblock ``Scalable magnetic field slam in 3d using gaussian process maps,''
\newblock in {\em 2018 21st International Conference on Information Fusion (FUSION)}, 2018, pp. 1353--1360.

\bibitem{footSLAM}
Patrick Robertson, Martin Frassl, Michael Angermann, Marek Doniec, Brian~J. Julian, Maria Garcia~Puyol, Mohammed Khider, Michael Lichtenstern, and Luigi Bruno,
\newblock ``Simultaneous localization and mapping for pedestrians using distortions of the local magnetic field intensity in large indoor environments,''
\newblock in {\em International Conference on Indoor Positioning and Indoor Navigation}, 2013, pp. 1--10.

\bibitem{vmag}
Zhenguang Liu, Luming Zhang, Qi~Liu, Yifang Yin, Li~Cheng, and Roger Zimmermann,
\newblock ``Fusion of magnetic and visual sensors for indoor localization: Infrastructure-free and more effective,''
\newblock {\em IEEE Transactions on Multimedia}, vol. 19, no. 4, pp. 874--888, 2017.

\bibitem{ogSIFT}
David~G Lowe,
\newblock ``Object recognition from local scale-invariant features,''
\newblock in {\em Proceedings of the seventh IEEE international conference on computer vision}. Ieee, 1999, vol.~2, pp. 1150--1157.

\bibitem{hog}
R~K McConnell,
\newblock ``Method of and apparatus for pattern recognition,''
\newblock 1 1986.

\bibitem{millane2020freetures}
Alexander Millane, Helen Oleynikova, Christian Lanegger, Jeffrey Delmerico, Juan Nieto, Roland Siegwart, Marc Pollefeys, and Cesar Cadena,
\newblock ``Freetures: Localization in signed distance function maps,'' arXiv, October 2020.

\bibitem{fpfh}
Radu~Bogdan Rusu, Nico Blodow, and Michael Beetz,
\newblock ``Fast point feature histograms (fpfh) for 3d registration,''
\newblock in {\em 2009 IEEE International Conference on Robotics and Automation}, 2009, pp. 3212--3217.

\bibitem{shot}
Samuele Salti, Federico Tombari, and Luigi {Di Stefano},
\newblock ``Shot: Unique signatures of histograms for surface and texture description,''
\newblock {\em Computer Vision and Image Understanding}, vol. 125, pp. 251--264, 2014.

\bibitem{linopsconstrained}
Carl Jidling, Niklas Wahlstr\"{o}m, Adrian Wills, and Thomas~B Sch\"{o}n,
\newblock ``Linearly constrained gaussian processes,''
\newblock in {\em Advances in Neural Information Processing Systems}, I.~Guyon, U.~Von Luxburg, S.~Bengio, H.~Wallach, R.~Fergus, S.~Vishwanathan, and R.~Garnett, Eds. 2017, vol.~30, Curran Associates, Inc.

\bibitem{rGP}
Manuel Schürch, Dario Azzimonti, Alessio Benavoli, and Marco Zaffalon,
\newblock ``Recursive estimation for sparse gaussian process regression,''
\newblock {\em Automatica}, vol. 120, pp. 109127, Oct. 2020.

\bibitem{sparseGPs}
Joaquin Qui\~{n}onero Candela and Carl~Edward Rasmussen,
\newblock ``A unifying view of sparse approximate gaussian process regression,''
\newblock {\em J. Mach. Learn. Res.}, vol. 6, pp. 1939–1959, Dec. 2005.

\bibitem{msac}
P.H.S. Torr and A.~Zisserman,
\newblock ``Mlesac: A new robust estimator with application to estimating image geometry,''
\newblock {\em Computer Vision and Image Understanding}, vol. 78, no. 1, pp. 138--156, 2000.

\bibitem{legentil2023calib}
Cedric Le~Gentil and Teresa Vidal-Calleja,
\newblock ``{A Gaussian Process approach for IMU to Pose Spatiotemporal Calibration},''
\newblock {\em Australasian Conference on Robotics and Automation, ACRA}, 2023.

\bibitem{magposegraphslam}
Jongdae Jung, Taekjun Oh, and Hyun Myung,
\newblock ``Magnetic field constraints and sequence-based matching for indoor pose graph slam,''
\newblock {\em Robotics and Autonomous Systems}, vol. 70, pp. 92--105, 2015.

\end{thebibliography}


\end{document}